\documentclass[conference]{IEEEtran}
\IEEEoverridecommandlockouts
\usepackage{cite}
\usepackage{amsmath,amssymb,amsfonts}
\usepackage{algorithmic}
\usepackage{graphicx}
\usepackage{textcomp}
\usepackage{xcolor}
\usepackage{booktabs}
\usepackage{siunitx} 

\usepackage{listings}
\usepackage{xcolor} 

\def\BibTeX{{\rm B\kern-.05em{\sc i\kern-.025em b}\kern-.08em
    T\kern-.1667em\lower.7ex\hbox{E}\kern-.125emX}}
\begin{document}

\title{Large Language Models for Explainable Threat Intelligence}

\author{\IEEEauthorblockN{Tiago Dinis$^1$ \ \ \ \
Roger Tavares$^2$ \ \ \ \
Miguel Correia$^1$}
\IEEEauthorblockA{\textit{$^1$INESC-ID, Instituto Superior Técnico, Universidade de Lisboa - Lisboa, Portugal} \\
\textit{$^2$UON - Lisboa, Portugal}
}}

\maketitle

\begin{abstract}
As cyber threats continue to grow in complexity, traditional security mechanisms struggle to keep up. Large language models (LLMs) offer significant potential in cybersecurity due to their advanced capabilities in text processing and generation. This paper explores the use of LLMs with retrieval-augmented generation (RAG) to obtain threat intelligence by combining real-time information retrieval with domain-specific data. The proposed system, RAGRecon, uses a LLM with RAG to answer questions about cybersecurity threats. Moreover, it makes this form of Artificial Intelligence (AI) explainable by generating and visually presenting to the user a knowledge graph for every reply. 
This increases the transparency and interpretability of the reasoning of the model, allowing analysts to better understand the connections 
made by the system based on the context recovered by the RAG system. 
We evaluated RAGRecon experimentally with two datasets and seven different LLMs and the responses matched the reference responses more than 91\% of the time for the best combinations. 
\end{abstract}

\begin{IEEEkeywords}
Large Language Models, Threat Intelligence, Retrieval-Augmented Generation, Natural Language Processing, Blockchain, Explainable AI
\end{IEEEkeywords}

\section{Introduction}

The volume and complexity of cyber threats continue to expand, driving organizations to rely on new technologies to provide more robust Cyber Threat Intelligence (CTI). Conventional security measures are often unable to handle and keep pace with the new range of cyber-attacks since the attackers adapt their methods. Large Language Models (LLMs) are highly effective in operating on textual data because of their ability to handle and process large unstructured semantic information. They demonstrate this by generating new semantic content and by analyzing massive volumes of previously collected data \cite{minaee2024large}\cite{naveed2023comprehensive}. Nevertheless, their capabilities can be further enhanced with a technique known as Retrieval-Augmented Generation (RAG), which combines real-time information retrieval with the generative capabilities of LLMs \cite{lewis2020retrieval}. Generative AI, as in LLMs, refers to AI systems designed to create original content, such as text, leveraging their training on vast datasets to generate coherent and contextually appropriate outputs \cite{goodfellow2020generative}. 

LLMs, such as the Generative Pre-trained Transformer (GPT) \cite{achiam2023gpt}, show impressive generative capabilities and are powerful tools for analyzing and interpreting large amounts of data. However, they use data processed during a slow and expensive training phase, so they cannot handle  
real-time information and specialized domains. RAG is a solution to this limitation. RAG connects LLMs with a retrieval mechanism that fetches documents from domain-specific datasets or live sources. This combination helps to ensure more accurate responses, a requirement in our fast-moving cybersecurity field. 

Utilizing RAG, organizations can deploy and benefit from pre-trained LLMs more effectively as they do not need to build or train models themselves, which is an unreasonable expense and requires a significant amount of computational resources. Instead, models can be quickly applied to cybersecurity problems by simply integrating them with security-specific data and reports and as a result extending the parametric memory of the model with non-parametric memory, assisting security analysts and CTI teams to assess threats with similar speed, create responses and produce real-time CTI. 

This paper presents \emph{RAGRecon}, a system aimed at providing CTI to cybersecurity personnel, e.g., to Security Operations Center (SOC) operators. 
RAGRecon uses a LLM with RAG to answer questions about cybersecurity threats. Moreover, it makes this form of AI explainable by generating and visually presenting to the user a Knowledge Graph (KG) for every reply. 
More in detail, RAGRecon retrieves information, called context, for a question asked by the user. The context is extracted from user-supplied CTI documents (e.g., PDF security reports) using RAG.
RAGRecon then prompts the LLM to identify relationships within that context and parses the information into a specific format. Using this, it creates a KG and answers the question based on the provided context. RAGRecon is designed to enable analysts to process and visualize connections in CTI data, facilitating the detection of vulnerabilities, the identification of anomalies, and the automation of responses to cyber threats, which is also applicable to the blockchain domain.

This project will leverage a wide spectrum of CTI reports, covering a far-reaching scope of cyber threats, from malware and phishing to advanced persistent threats. Moreover, the project will place particular emphasis on the domain of blockchain technology. As blockchain technology continues to acquire recognition across industries, it has become an increasingly attractive target for cybercriminals \cite{hornuf2023cybercrime}. By focusing on strengthening security within the blockchain domain, this project strives to enhance our abilities to identify and respond to particular threats to this area, such as flaws within smart contracts, hacks of cryptocurrency wallets, and exploits affecting decentralized financial applications\cite{cong2023blockchain}. 

The motivation for RAGRecon comes from the need to address the complexities of analyzing and responding to cyber threats in an increasingly data-intensive landscape. Employing LLMs allows the system to efficiently parse and process natural language CTI with accuracy and adaptability. The integration of RAG facilitates the retrieval of the most relevant and contextually accurate information, addressing the challenges of information overload while reducing the occurrence of hallucinations in answers. RAGRecon includes an explainable AI component that synthesizes relevant context to answer user questions. This benefits users by simplifying complex information while also highlighting which parts of the context the language model found most relevant to generate its response. The use of explainable AI ensures that the system's reasoning is transparent, fostering trust and enabling users to understand the rationale behind its answers. Finally, the incorporation of a KG provides an intuitive visual representation of the relevant interconnected CTI data within the user's documents necessary to answer the question.

The experimental evaluation, conducted on two custom-built datasets\footnote{Available at: https://github.com/Tiago-Din/RAGRecon-Datasets}, one for conventional CTI (i.e., CTI about common security threats) based on 24 PDF reports, and one for Blockchain CTI based on 28 PDF reports, each comprising 50 questions. 
The system achieved high Faithfulness scores, consistently exceeding 0.8 out of 1.0, indicating a low rate of factual hallucinations. Furthermore, it exhibited efficient Context Relevance, using approximately 8\% of the retrieved context on average to generate answers. To validate the LLM self-evaluation methodology, a manual analysis of 2,050 automated decisions was performed, confirming the high accuracy of the approach with correct decision rates generally ranging from 90\% to 97\% with 7 LLMs. This verification also substantiated that minor performance variations resulted from occasional errors by both the generation model and the self-evaluation model. A slight performance advantage was observed with the blockchain dataset, although it was considered likely to be attributable to the limited dataset size rather than to a statistically significant trend.

The paper has the following main contributions: 1) RAGRecon, a system that uses LLMs, RAG, and KGs to provide explainable CTI; 2) an experimental evaluation of RAGRecon with two CTI datasets and 7 different LLMs; 3) insights into performance variations between the two datasets and the different LLMs used.

\section{Background and Related Work}

This section explores three fundamental components central to our proposal: LLMs, RAG, and CTI. 

\subsection{Large Language Models}

LLMs are sophisticated AI systems designed to process and generate human language. Built on deep learning architectures like the transformer, they are trained on vast and diverse datasets, enabling them to perform tasks such as text completion, translation, and question-answering. The scale of these models, often containing billions of parameters, allows them to capture complex linguistic patterns. However, LLMs have inherent limitations, including knowledge cutoffs, a tendency to ``hallucinate" (generate plausible-sounding but factually incorrect information) and potential biases inherited from their training data \cite{roffo2024exploring, pal2023med, dahl2024large}.

Architecturally, LLMs can be distinguished by their core components. Encoder-only models like BERT (Bidirectional Encoder Representations from Transformers) \cite{devlin2019bert} are designed to understand and represent input text, excelling at tasks like classification and entity recognition. Decoder-only models like GPT focus on generating text sequences \cite{achiam2023gpt}. Encoder-decoder models such as T5 (Text-to-Text Transfer Transformer) combine both functions to transform input sequences into new output sequences, making them suitable for summarization and translation \cite{minaee2024large, yenduri2024gpt}.

LLMs can be classified as public or private. Public models, such as Mistral 7B or those available through APIs like GPT-4o, democratize access to AI but raise concerns about misuse and security. Private models, such as Google's PaLM, offer greater control and data security for organizations, but require significant resources to develop and maintain.

To address the black-box nature of these models, the field of Explainable AI (XAI) aims to make their decision-making processes transparent and interpretable \cite{arrieta2020explainable}. Graphs are a particularly effective tool in XAI, helping to visualize data flow and relationships within the model, which is crucial for building trust and ensuring accountability \cite{schneider2024explainable, rajabi2022knowledge}.

\subsection{Retrieval-Augmented Generation}

RAG is a framework that enhances LLMs by connecting them to external knowledge sources in real time. Introduced by Lewis et al.\ \cite{lewis2020retrieval}, RAG combines the parametric knowledge stored within an LLM's parameters with non-parametric knowledge retrieved from a vector store. This hybrid approach allows the model to access up-to-date, factual information, significantly reducing hallucinations and improving the accuracy of its responses. A key advantage of RAG is that the external knowledge base can be updated without retraining the entire model. The RAG process involves two main steps: retrieval and generation.

\begin{enumerate}
  \item Retrieval: When a query is received, the system first retrieves relevant documents from a knowledge base. This can be done using two main methods:
sparse retrieval, based on keyword matching, and dense retrieval, which uses embeddings to represent the semantic meaning of text. Dense retrieval converts both the query and the documents into vectors and finds the most relevant documents based on vector similarity. Dense retrieval is better at understanding context, but is more computationally expensive.
  \item Generation: The retrieved information is then passed to the LLM along with the original query and the model generates a response based on this augmented context.
\end{enumerate}

There are two primary RAG variants \cite{lewis2020retrieval}:

\begin{itemize}
      \item RAG-Sequence: Retrieves a set of documents once and uses them to generate the entire response in a single pass. It is effective for tasks requiring a broad, coherent answer synthesized from multiple sources.
      \item RAG-Token: Retrieves information dynamically for each token being generated. This allows the model to adjust its focus in real-time based on the evolving context, making it suitable for more dynamic responses.
\end{itemize}

RAG is central to modern Question Answering (QA) systems. These systems can be extractive (pulling exact answers from text) or abstractive (generating new, summarized answers). They can also be closed domain (specialized in one topic) or open domain (covering a wide range of topics). RAG architecture is highly adaptable to these types of QA systems.

\subsection{Cyber Threat Intelligence}

Threat intelligence is the process of gathering, analyzing and acting upon data about cyber threats to improve cybersecurity defenses. It helps organizations understand adversary tactics and make informed security decisions. CTI is often categorized into two types:

\begin{itemize}
      \item Operational CTI: Focuses on immediate, actionable threats, providing technical details like Indicators of Compromise (IOCs) to help security teams detect and respond to ongoing attacks.
      \item Strategic CTI: Provides a high-level view of the threat landscape, helping decision makers prioritize resources and anticipate future challenges.
\end{itemize}

A significant challenge in CTI is the manual effort required to extract structured information from unstructured reports, a process that can be extremely time-consuming \cite{siracusano2023time}. Automating this process is crucial for making CTI faster and more effective. Researchers have proposed systems to automate the classification and enrichment of threat data from various sources, including Open Source Intelligence (OSINT) from platforms such as Twitter \cite{martins2022generating, alves2021processing}.

LLMs have shown promise in automating CTI tasks. Shafee et al.\ \cite{shafee2024evaluation} found that LLMs perform well in binary classification of threat data, but struggle with more complex tasks like Named Entity Recognition (NER), which is essential to extract specific entities such as malware names or threat actor groups. This highlights a key area for improvement.

\subsection{Enhancing CTI with LLMs and RAG}

The integration of RAG with LLMs offers a powerful solution to the challenges in CTI \cite{zhang2024llms}. By providing LLMs with real-time access to specialized databases (e.g., threat reports, vulnerability databases), RAG enables the generation of high-quality, contextualized CTI \cite{rajapaksha2024rag}.

This approach directly addresses the needs of SOC analysts, who often must manually contextualize global CTI for their specific organization. Mitra et al.\ \cite{mitra2024localintel} introduced LOCALINTEL, a RAG-based system that automates this process. LOCALINTEL retrieves information from global sources like the National Vulnerability Database (NVD) and combines it with an organization's internal knowledge base to produce tailored, actionable intelligence. This system significantly reduces manual labor and improves the accuracy of local threat assessments.

Furthermore, RAG-enhanced LLMs can improve the understanding of adversary Tactics, Techniques and Procedures (TTPs), which are cataloged in frameworks like MITRE ATT\&CK. Fayyazi et al.\ \cite{fayyazi2023advancing} found that while large decoder-only models like GPT-3.5 can struggle with precision when analyzing TTPs, integrating a RAG component significantly boosts their performance. This demonstrates the potential of combining LLMs with RAG to interpret complex cyberattack behaviors.

Our work contributes to the state-of-the-art by presenting a novel method for visualizing an LLM's reasoning using KGs. Our system synthesizes context retrieved via RAG, prioritizing information based on its relevance to the user's question. Then it generates a visual map of key concepts and their interconnections, providing users with clear visibility into the data points that support the answer.

\section{RAGRecon}

This section presents RAGRecon. 
The section focuses on the approach; additional details are left for Section \ref{sec:implementation} that presents its implementation.

\subsection{Architecture}

\begin{figure}[th]  
\centering
\includegraphics[width=1\columnwidth]{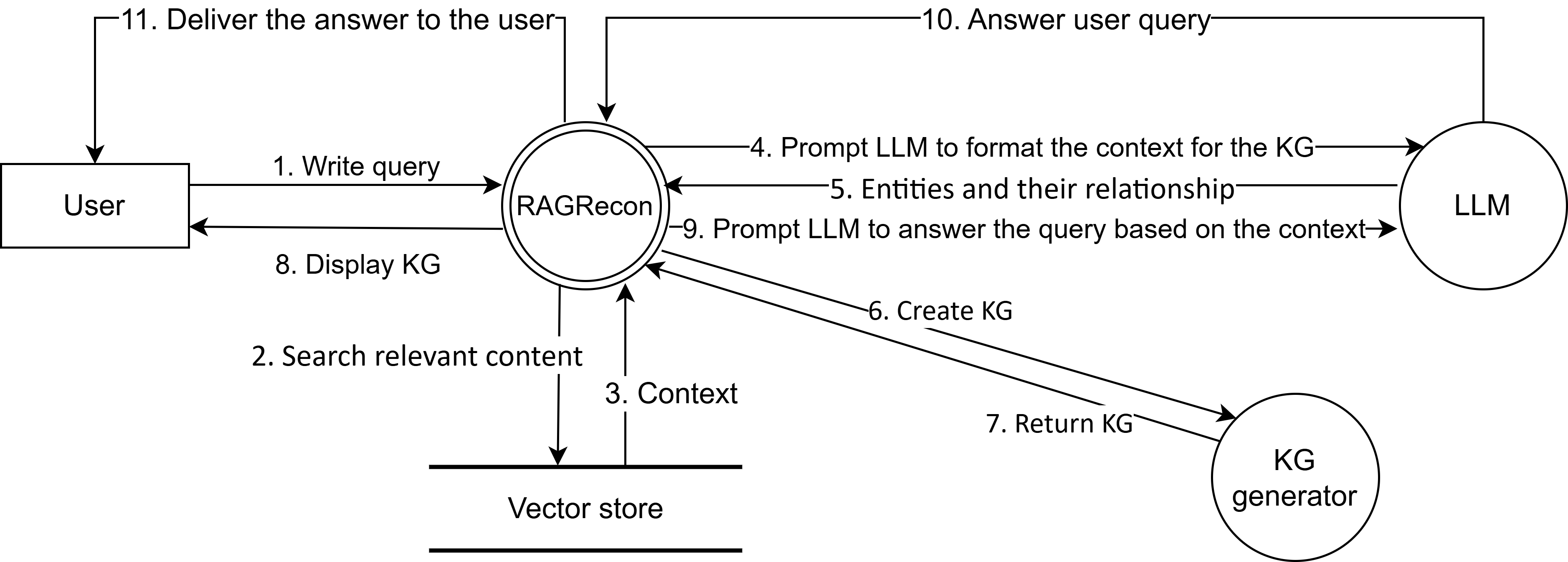}  
\caption{Data flow diagram of the proposed system architecture}
\label{fig:data_flow_system}
\end{figure}

The architecture of the system is shown in Figure \ref{fig:data_flow_system}. The central component, designated RAGRecon in the figure, essentially processes user queries. 

This architecture leverages RAG to extract and structure relevant information from the vector store, delivering accurate and context-aware responses to user queries. The process begins when the user submits a query via the command-line interface. Upon receiving the query, RAGRecon searches the vector store, which is preloaded with relevant data sources (in the form of embeddings, as explained later), to retrieve information that is contextually aligned with the user's input.

RAGRecon retrieves the context and processes it to construct the KG and guide the LLM's tasks. RAGRecon prompts the LLM to format this extracted context into a structured format, which is then used to build the KG. This graph visually and logically represents the domain knowledge context relevant to the query.

Once the KG is constructed, RAGRecon calls again the LLM to interpret the query within the previously retrieved context. Using contextual information, the LLM generates a contextually rich answer to the user's query. Finally, the KG and the generated answer are delivered to the user, completing the process.

\subsection{Data Ingestion and Vectorization}

The preload of the documents into the system initiates a pipeline that processes and prepares a collection of documents for search and retrieval of similarity. Initially, the documents are loaded and then split into smaller, manageable chunks. This step is critical, particularly for handling long documents \cite{liu2023lostmiddlelanguagemodels}, as it ensures that each chunk is appropriately sized for embedding models, allowing for effective processing without overwhelming the system or losing important context.

After splitting the documents, RAGRecon proceeds to generate vector embeddings for each chunk using a pre-trained model. These embeddings are high-dimensional numerical representations that capture the semantic meaning of the text. Rather than relying on keyword matching, the model translates each chunk of text into a vector that encodes its contextual information, meaning that even if the exact phrasing does not match a query, similar content can still be retrieved based on its meaning.

Once the embeddings are generated, they are stored in a vector store, a specialized database able to store and index high-dimensional vectors. Through these embeddings, the system can identify and retrieve the most relevant document chunks based on their semantic similarity to the user's query, rather than simply matching words or phrases.

\subsection{Interactive RAG}

The system's core functionality is a dual-output architecture that processes user queries to generate both a detailed textual response and a visual KG. Operating in an interactive loop similar to a conversational chatbot, the workflow enables a dynamic and exploratory user experience. This real-time iterative cycle allows users to ask follow-up questions, leveraging both textual and visual summaries to build a comprehensive understanding.

\subsubsection{Textual Response Generation}

Upon receiving a user query, the system's primary objective is to generate an informed and coherent textual answer. This process is initiated by retrieving the relevant context from the vector store, which contains precomputed vector embeddings of document chunks. These high-dimensional vectors encode the semantic meaning of the text, enabling efficient and accurate information retrieval.
The system performs a similarity search against the vector store to identify the top-K document chunks most relevant to the user's query. Following the RAG-Sequence approach, these retrieved chunks are concatenated into a single block of text. This consolidated context is then integrated with the original user question into a RAG prompt. The prompt is engineered to instruct the LLM to synthesize the information and generate a user-friendly explanation while maintaining a natural conversational tone. Finally, the crafted prompt is passed to the LLM, which returns a detailed, contextually-grounded natural language response.

\subsubsection{Knowledge Graph Generation and Visualization}

To complement the textual response, the system generates and visualizes the contextual information as a Knowledge Graph (KG), providing users with a clear summary of the relationships between key concepts. 
Using the same retrieved context chunks, the system formulates a separate, graph-related prompt. This prompt specifically instructs the LLM to analyze the text and extract the primary entities (nodes) and the relationships (edges) that connect them. The LLM generates a structured textual description of these relationships. This output is then processed to construct the KG, where each entity is represented as a node, and each relationship is depicted as a connecting edge. This visualization provides an intuitive map of the information, helping users grasp complex connections at a glance.

\section{Implementation of RAGRecon}
\label{sec:implementation}

\begin{figure*}[th]  
\centering
\includegraphics[width=2\columnwidth]{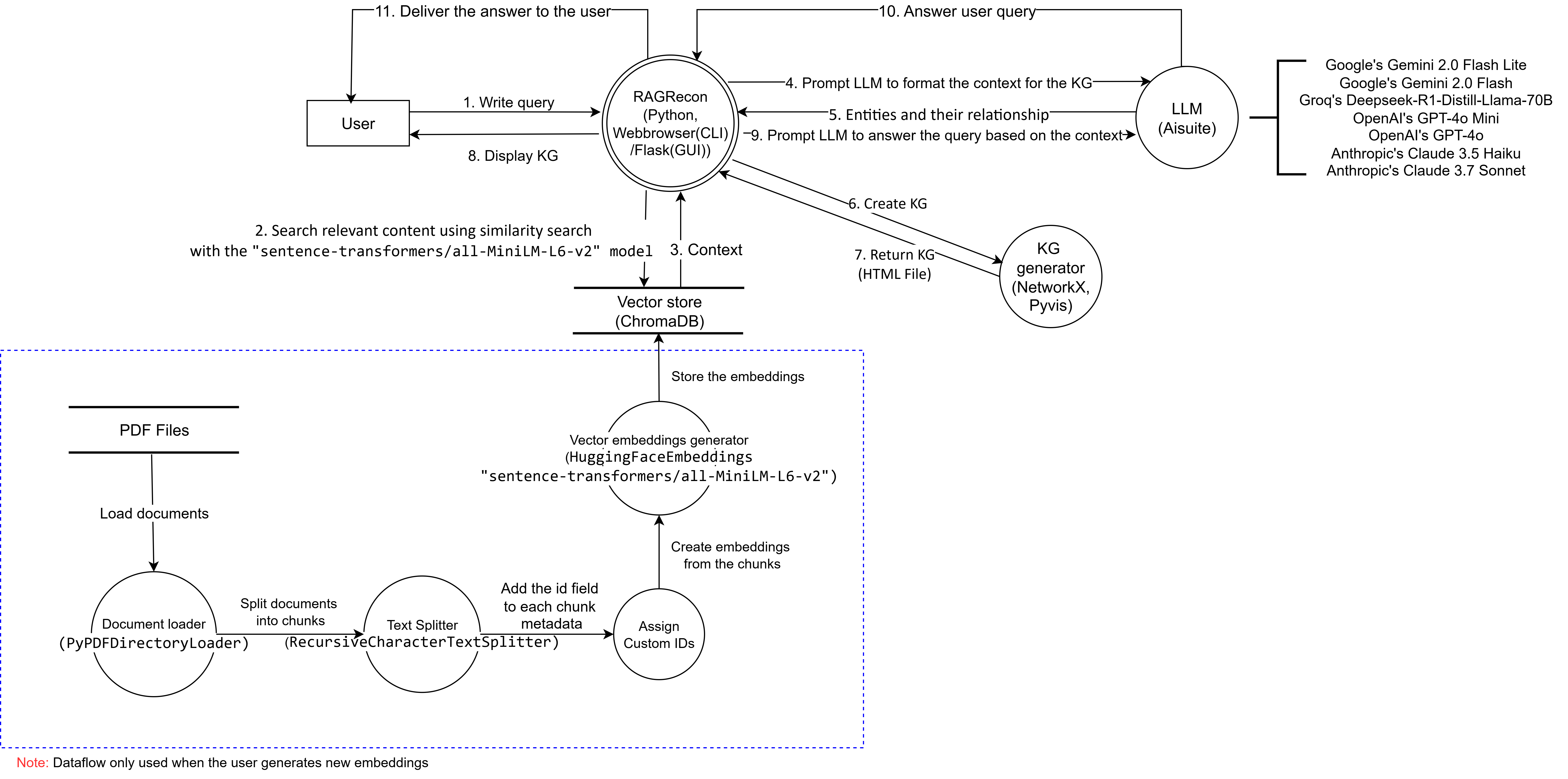}  
\caption{Data flow diagram with the technologies used in the architecture}
\label{fig:data_flow_system_tech}
\end{figure*}

This section details the technical implementation of the RAGRecon system, represented in Figure \ref{fig:data_flow_system_tech}. 
The system is built with a modular architecture, featuring a back-end for data processing and a choice of two distinct user interfaces: a script-based command line interface (CLI) and an interactive graphic user interface (GUI).

The implementation leverages the LangChain framework\footnote{https://www.langchain.com/langchain} to orchestrate the retrieval process, from data ingestion and vectorization to real-time query handling and context retrieval, ChromaDB\footnote{https://docs.trychroma.com/docs/overview/introduction} a DBMS for vector storage also targeted at LLMs, the Flask\footnote{https://flask.palletsprojects.com/en/stable/} framework for the web interface and various LLMs via aisuite\footnote{https://github.com/andrewyng/aisuite}.

The foundational logic of RAGRecon is divided into two primary stages: a processing pipeline to create a queryable knowledge base and an interactive pipeline to process queries, retrieve context and generate responses. This core logic is common to the CLI and the GUI.

\subsection{Data Ingestion and Vectorization}

The initial phase of the system involves processing the source PDF documents and converting them into a queryable vector format. This is handled by a dedicated data preparation pipeline that performs a series of sequential steps.

\subsubsection{Document Loading and Pre-processing}

Source documents are placed in a designated data directory. The system utilizes the PyPDFDirectoryLoader from LangChain to load all PDF files from this location. This loader treats each page of every PDF as a distinct Document object, preserving its content and metadata, such as the source filename and page number.

\subsubsection{Text Chunking}

To ensure that the text segments are manageable in size for the embedding model and fit within the context windows of the LLM, the loaded documents are split into smaller chunks. This process uses LangChain's RecursiveCharacterTextSplitter configured with the following parameters:

\begin{itemize}
  \item Chunk Size: 1000 characters.
  
  \item Chunk Overlap: 100 characters. This overlap helps maintain semantic context between adjacent chunks.
  
  \item Separators: A hierarchical list of separators is used to split text along logical boundaries, starting with paragraphs and progressively moving to smaller units:

  \small{\verb|(["\n\n", "\n", "(?<=\. )", " ", ""])|}
\end{itemize}

\subsubsection{Custom Metadata and ID Generation}

A crucial step in this phase is the assignment of a unique identifier to each text chunk. After splitting, the script iterates through each chunk and enriches its metadata with a custom id. This ID has the format:
\texttt{\small{filename\_p\textless page\_number\textgreater\_c\textless chunk\_index\_on
\_that\_page\textgreater}}
For example, the third chunk from the fifth page (0-indexed page 4) of a file named report.pdf would receive the ID \texttt{\small{report.pdf\_p\textless 4\textgreater\_c\textless 2\textgreater}}. This identification scheme is essential for later retrieving and reconstructing the context of a document in its original order.

\subsubsection{Embedding Generation and Vector Storage}

The processed text chunks are then converted into numerical representations (embeddings). This is accomplished using LangChain's HuggingFaceEmbeddings wrapper, which loads the sentence-transformers/all-MiniLM-L6-v2 model. This model is chosen as it is free to use, efficient and effective in generating meaningful semantic vectors for sentences and paragraphs.
The resulting embeddings, along with their corresponding text content and metadata (including the custom ID), are stored in a ChromaDB vector store. The database is configured to persist on disk in the ./chroma\_db directory. To ensure a fresh build during each execution, the script automatically deletes any pre-existing database directory before creating a new one.

\subsection{Interactive RAG and Knowledge Graph Generation}

The second part of the system is the interactive query engine, which provides an interface for users to query the indexed documents. This engine manages context retrieval, LLM interaction and visualization features.

\subsubsection{Query Processing and Context Retrieval}

When a user submits a query, the following steps occur:

\begin{itemize}
  \item Query Embedding: The user's query is converted into an embedding using the same sentence-transformers/all-MiniLM-L6-v2 model.
  \item Similarity Search: The system performs a similarity search against the ChromaDB vector store. Retrieves the k=6 text chunks most semantically similar to the query embedding.
  \item Context Assembly: The content of these 6 chunks is concatenated to form the context that will be fed to the LLM.
\end{itemize}

\subsubsection{LLM Integration and Prompting}

The RAGRecon system is designed to be model-agnostic, supporting a variety of LLMs from providers like Google, OpenAI, Anthropic and Groq. A carefully crafted prompt is generated that combines the user's original query with the retrieved context. The prompt includes system instructions that guide the LLM to act as a helpful bot, answer in a conversational tone, and base its response on the provided context. The system also maintains a conversation history that is formatted in a model-agnostic way. This allows a user to switch between different LLMs in the same conversation while still enabling the active model to use previous turns for more coherent, multi-turn dialogues.

\subsubsection{Knowledge Graph Visualization}

A key feature of RAGRecon is its ability to generate and visualize knowledge graphs to represent relationships within the text. The process is as follows:

\begin{itemize}
  \item Graph Data Extraction: A specialized prompt instructs the selected LLM to analyze the retrieved context and extract entities and their relationships. The LLM is required to return this information in a structured JSON format, like [{``subject": ``A", ``relationship": ``B", ``object": ``C"}], where the letters correspond to the pieces of relevant information identified by the LLM.
  \item Visualization: This is handled by a function that parses the LLM-generated JSON and builds a graph using NetworkX\footnote{https://networkx.org/documentation/stable/index.html} and the Pyvis\footnote{https://pyvis.readthedocs.io/en/latest/index.html} Python libraries to create an interactive HTML file.
\end{itemize}

\subsubsection{Example}

To illustrate the system's capabilities, this section provides a concrete example of how an input query is processed to generate a response and a corresponding knowledge graph.

\paragraph{Input Query}

Let us assume a user submits the following query to the RAGRecon system using the anthropic provider with the claude-3-7-sonnet-20250219 model:

\noindent\emph{\small{``What is Broken Object Level Authorization (BOLA)?"}}

\paragraph{Context Retrieval}

The system converts this query into an embedding and performs a similarity search. The top k=6 retrieved chunks are concatenated to form the following context (a condensed version for clarity):

\noindent\emph{\small{``(Insecure Direct Object Reference). BOLA arises from APIs exposing object identifiers 
through their endpoints, posing significant Object Level Access Control concerns. 
This vulnerability allows attackers to manipulate or access API ..."}}

\paragraph{LLM-Generated Knowledge Graph Data}

The system then provides the retrieved context to the LLM with a specialized prompt, instructing it to extract entities and relationships in a structured JSON format. The LLM returns the following data:

{\footnotesize\noindent[{``subject": ``BOLA", ``relationship": ``arises from", ``object": ``APIs exposing object identifiers"}, {``subject": ``BOLA", ``relationship": ``poses", ``object": ``Object Level Access Control concerns"}, {``subject": ``BOLA", ``relationship": ``allows", ``object": ``attackers to manipulate or access API data/resources without proper authorization"}, {``subject": ``BOLA", ``relationship": ``leads to", ``object": ``severe consequences"}, {``subject": ``BOLA vulnerabilities", ``relationship": ``can result in", ``object": ``unauthorized access, breaches, and misuse of critical functionalities"}, {``subject": ``security teams", ``relationship": ``should implement", ``object": ``robust monitoring and logging mechanisms"}, {``subject": ``BOLA", ``relationship": ``is also known as", ``object": ``IDOR"}, {``subject": ``OWASP API Security Top 10", ``relationship": ``was updated in", ``object": ``2023"}, {``subject": ``OWASP API Security Top 10", ``relationship": ``is developed by", ``object": ``Open Worldwide Application Security Project"}, {``subject": ``BOLA", ``relationship": ``is top priority on", ``object": ``the list"}]}

\paragraph{Resulting Knowledge Graph Visualization}

\begin{figure}[th] 
\centering
\includegraphics[width=1\columnwidth]{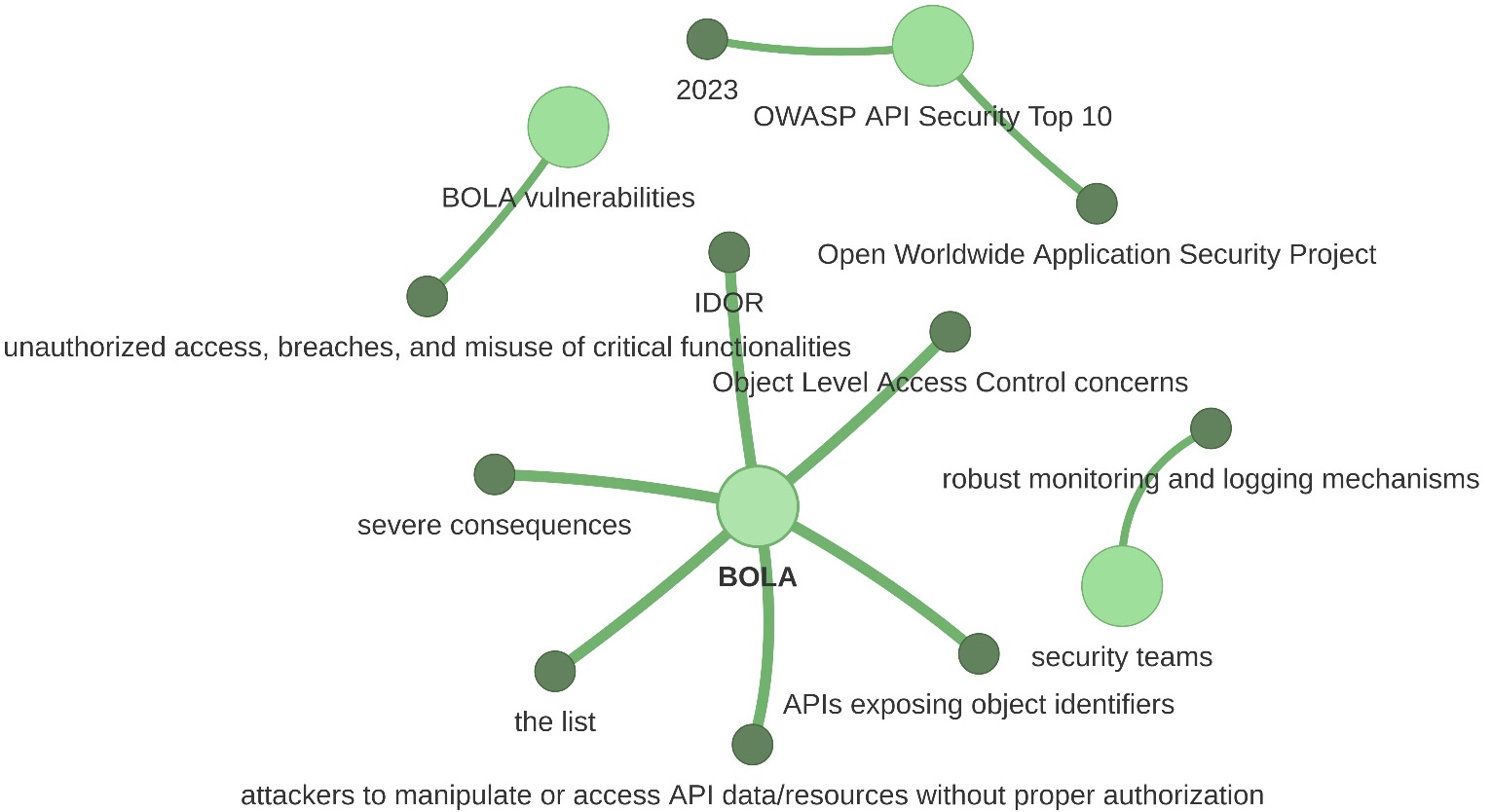}  
\caption{Knowledge Graph Example}
\label{fig:graph}
\end{figure}

The JSON output is then parsed to build a graph using NetworkX, which is then rendered as an interactive HTML file with Pyvis (Figure \ref{fig:graph}).

The KG is a tool to help users comprehend large and complex contexts passed to an LLM. When faced with a long document, a user often encounters a ``wall of text" that leads to information overload. The linear nature of prose makes it difficult to quickly identify the main subjects and understand the relationships between concepts, often found pages apart, forcing the user to build a complex mental map from scratch in a cognitively demanding task.

The KG solves this problem by transforming unstructured text into a structured relational map that acts as a visual summary. It allows a user to grasp the main topics at a glance, as the most important concepts are represented by nodes. In Figure \ref{fig:graph}, the central subject is BOLA. However, the true power of this visualization is its ability to reveal the relationships between these concepts through its connecting lines. The user no longer needs to scan the entire document to understand how the ideas are linked. Observing the graph, 
one can instantly see these precise connections: the vulnerability BOLA arises from APIs exposing object identifiers and poses Object Level Access Control concerns. The graph clarifies that BOLA is also known as IDOR and 
shows that it allows attackers to manipulate or access API data/resources without proper authorization, which in turn leads to severe consequences. Furthermore, it connects other facts, visualizing that BOLA is top priority on ``the list," which it identifies as the OWASP API Security Top 10 that was updated in 2023.

In practice, this visualization acts as a factual map of the original context based on what the LLM found relevant to answer the question and as a help in formulating better questions. A user can review the graph to build an accurate mental model and then ask the LLM more specific questions based on the established relationships. This reduces cognitive load and aids the user to explore, verify, and question the LLM's knowledge base with much greater precision and speed.

\paragraph{Answer Generation}

The retrieved context and the original query are combined into a prompt. The LLM then generates the following conversational response based on this provided context:

\noindent\emph{\small{``Broken Object Level Authorization (BOLA), also known as Insecure Direct Object Reference (IDOR), is a security vulnerability that occurs when APIs expose object identifiers through their endpoints. This creates significant Object Level Access Control issues, allowing attackers to manipulate or access API data and resources without proper authorization.}}

\noindent\emph{\small{BOLA is the top priority on the OWASP API Security Top 10 list (updated in 2023). Although an average of 1.6 API endpoints per organization are at risk of BOLA abuse, the consequences can be severe, including unauthorized access, data breaches, and misuse of critical functionalities.}}

\noindent\emph{\small{To mitigate BOLA vulnerabilities, security teams should implement robust monitoring and logging mechanisms to track API usage, detect anomalies, and identify potential unauthorized access."}}

\subsection{User Interfaces}

RAGRecon provides two distinct interfaces that build on the core architecture, catering to different user preferences.

\subsubsection{Command-Line Interface}

The CLI provides a script-based way to interact with the RAGRecon system. It is designed for users who are comfortable working in a terminal environment.
The RAGRecon script encapsulates the interactive logic in a while True loop that continuously prompts the user for input. The user can select their desired LLM at the beginning of the session and then begin asking questions. In addition to standard queries, the CLI supports special flag-based commands for managing the session, such as /quit to exit, /help to display options, /clean\_history to reset the conversation, and /change\_model to switch LLMs mid-session. Each query triggers the full RAG pipeline and the generated KG HTML file is automatically opened in the user's default web browser.

\subsubsection{Graphical User Interface}

The GUI is web-based user interface built with the Flask web framework. It provides a more intuitive and interactive experience for managing documents, posing queries and visualizing results.
The GUI's backend is a Flask application that exposes several API endpoints (@app.route) that respond to HTTP requests from the user's browser. This event-driven model replaces the CLI's input loop. Key endpoints manage query submission, model selection, file uploads/deletions and the triggering of the embedding pipeline. The application state, such as the conversation history and the selected model, is managed via global variables within the Flask application.




\section{Experimental Evaluation}

The performance of the system was evaluated by comparing the answers it generated with predefined reference answers for a given set of questions. To automate this comparison, we used an LLM self-evaluation methodology introduced by Ren et al.\ in \cite{ren2023self}. This involves using a separate LLM to assess the semantic similarity between the reference answer and the answer generated by the RAGRecon system. We not only used this methodology for all answers, but also manually assessed all of them.

\subsection{Dataset Construction}

To facilitate the evaluation, two datasets were constructed:

\begin{enumerate}
    \item Conventional CTI: The source material for this analysis consisted of 24 PDF reports on CTI, produced by various companies in the field.
    
    \item Blockchain CTI: The process was repeated with 28 PDF reports specific to the blockchain domain.
\end{enumerate}

Each dataset was built as follows:

\begin{itemize}
    \item Question Generation: Our process involved prompting Gemini 2.5 Flash to generate a pool of questions for each document. We then manually selected 50 of these questions to form the datasets. For example, a question from the conventional CTI dataset asked \emph{``Which ransomware group targeted the UK's Royal Mail in January 2023?"} while a question from the blockchain dataset sought to \emph{``Explain the concept of a 51\% attack in the context of a permissionless distributed ledger."}
    
    \item Reference Answer Generation: The same LLM was then used to generate a reference answer for each question, using the source document as context.
    
    \item Manual Review: Each question and reference answer pair underwent a manual review to ensure accuracy and eliminate any instances of hallucination.
\end{itemize}

\subsection{Benchmarking Framework}

\begin{figure}[th] 
\centering
\includegraphics[width=1\columnwidth]{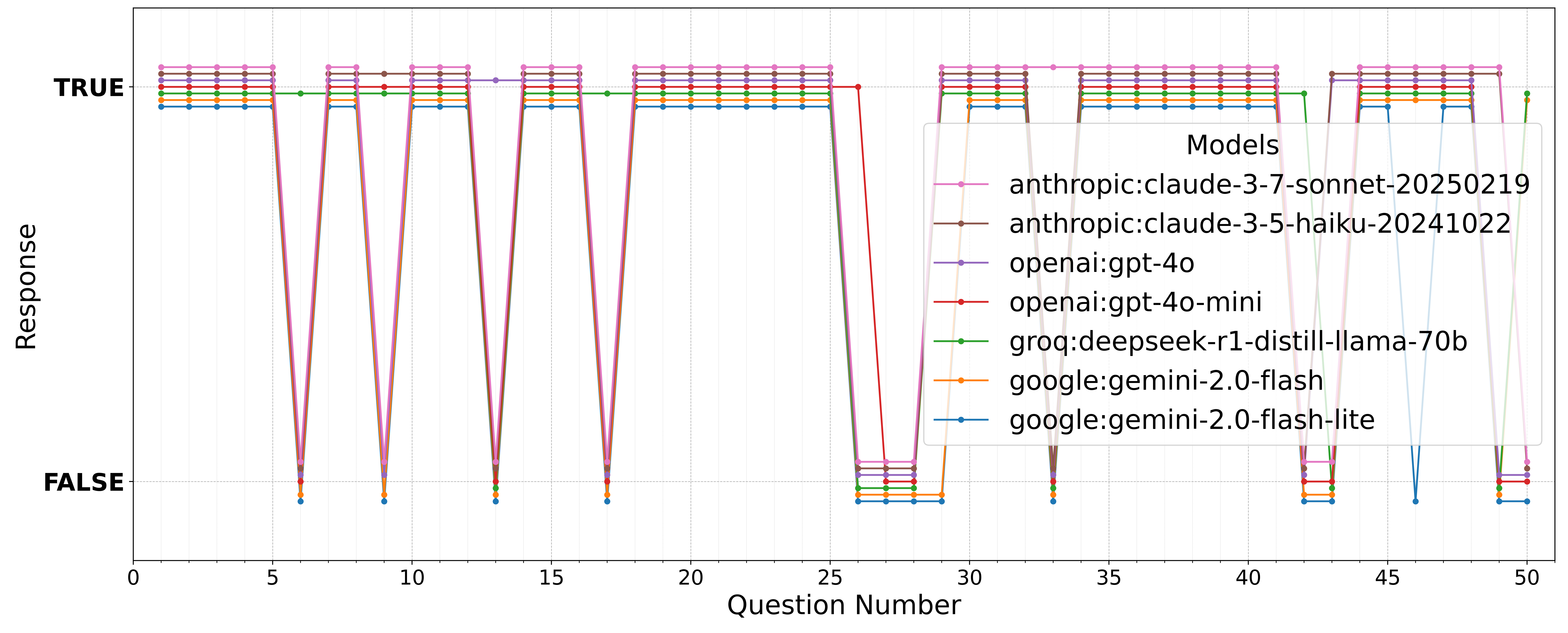}  
\caption{Comparison different models Round 3 Conventional CTI Dataset}
\label{fig:comparison_different_models_same_round_results_dataset-security_round_3}
\end{figure}

To benchmark performance, we evaluated seven distinct LLMs from four providers. The models included Google's Gemini 2.0 Flash and Flash Lite (accessed via the free tier), OpenAI's GPT-4o and GPT-4o Mini (paid access), Anthropic's Claude 3.7 Sonnet and Claude 3.5 Haiku (paid access) and Groq's freely available Deepseek-R1-Distill-Llama-70B. The performance of each LLM was assessed across multiple testing rounds on both the conventional security and the blockchain datasets. Visualizations, such as those exemplified in Figure \ref{fig:comparison_different_models_same_round_results_dataset-security_round_3}, were used to analyze and compare their performance.

\subsection{Key Evaluation Metrics}

Following the methodology outlined by Es et al. in \cite{es2024ragas}, the system was evaluated using two primary metrics:

\begin{itemize}
    \item Faithfulness: This metric measures the degree to which the generated answer is factually grounded in the provided context. High faithfulness ensures that the response is accurate, avoids hallucination, and is justified by the retrieved information.

    \begin{small}\em
    \[
    \text{Faithfulness} = \frac{\text{Number of Statements Supported by Context}}{\text{Total Number of Statements}}
    \]
    \end{small}

    How it is calculated:
    \begin{enumerate}
        \item An LLM breaks down the generated RAG answer into individual statements.
        \item A second LLM call verifies each statement against the context, producing a ``Yes" or ``No" verdict for each.
        \item Two more LLM calls are made to count the total number of ``Yes" verdicts and ``No" verdicts.
        \item The final score is the ratio of ``Yes" verdicts to the total number of statements (``Yes" + ``No").
    \end{enumerate}
    
    \item Context Relevance: This metric checks how much useful information is in the data the system retrieves. It is crucial for efficiency, as it ensures that the context provided to the LLM contains minimal irrelevant information, thereby reducing the computational cost associated with processing long passages.

    \begin{small}\em
    \[
    \text{Context Relevance} = \frac{\text{Number of Relevant Sentences}}{\text{Total Number of Sentences in Context}}
    \]
    \end{small}

    How it is calculated:
    \begin{enumerate}
        \item An LLM extracts sentences from the provided context that it deems relevant to the question.
        \item A second LLM call counts the number of these relevant sentences.
        \item A third LLM call counts the total number of sentences in the original context.
        \item The final score is the ratio of relevant sentences to the total sentences.
    \end{enumerate}
    
\end{itemize}

\subsection{Results and Analysis}

\subsubsection{System Pipeline and Determinism}

Log analysis provided key insights into the system's internal workings. These logs document the automated tests conducted on the RAGRecon system. The tests evaluate how effectively the benchmarked models answer questions from two distinct datasets. The logs detail each critical stage of the process, including:
\begin{itemize}
    \item Ingestion: Loading documents for embedding generation.
    \item Retrieval: Extracting relevant context from the vector store when RAGRecon is running.
    \item LLM self-evalution: Running the RAGRecon system and comparing its answer with the Reference Answer.
    \item Evaluation: The key performance metrics discussed in the previous section.
\end{itemize}

The logs show the pipeline is deterministic from ingestion to retrieval. For a given dataset, the chunking process consistently produces the same number of chunks with identical splits and metadata. Consequently, a given query always retrieves the exact same context as it will find the exact same vectors.

While the answers from the RAGRecon system are semantically consistent, their phrasing may change from run to run due to the nature of LLMs.

\subsubsection{Analysis of Retrieval and Generation Metrics (at Top-K=6)}

The evaluation highlighted strong and consistent performance in both context relevance and faithfulness.

Context Relevance: The system demonstrated efficient context utilization, using approximately 8\% of the retrieved information on average to formulate an answer.

Faithfulness: The faithfulness of the generated answers was consistently high for both datasets, indicating a low rate of model hallucination. With only a single minor exception in one round, the average faithfulness score consistently exceeded 0.8 out of 1.0.

\subsubsection{LLM Self Evaluation Manual Verification}

\begin{figure}[th] 
\centering
\includegraphics[width=1\columnwidth]{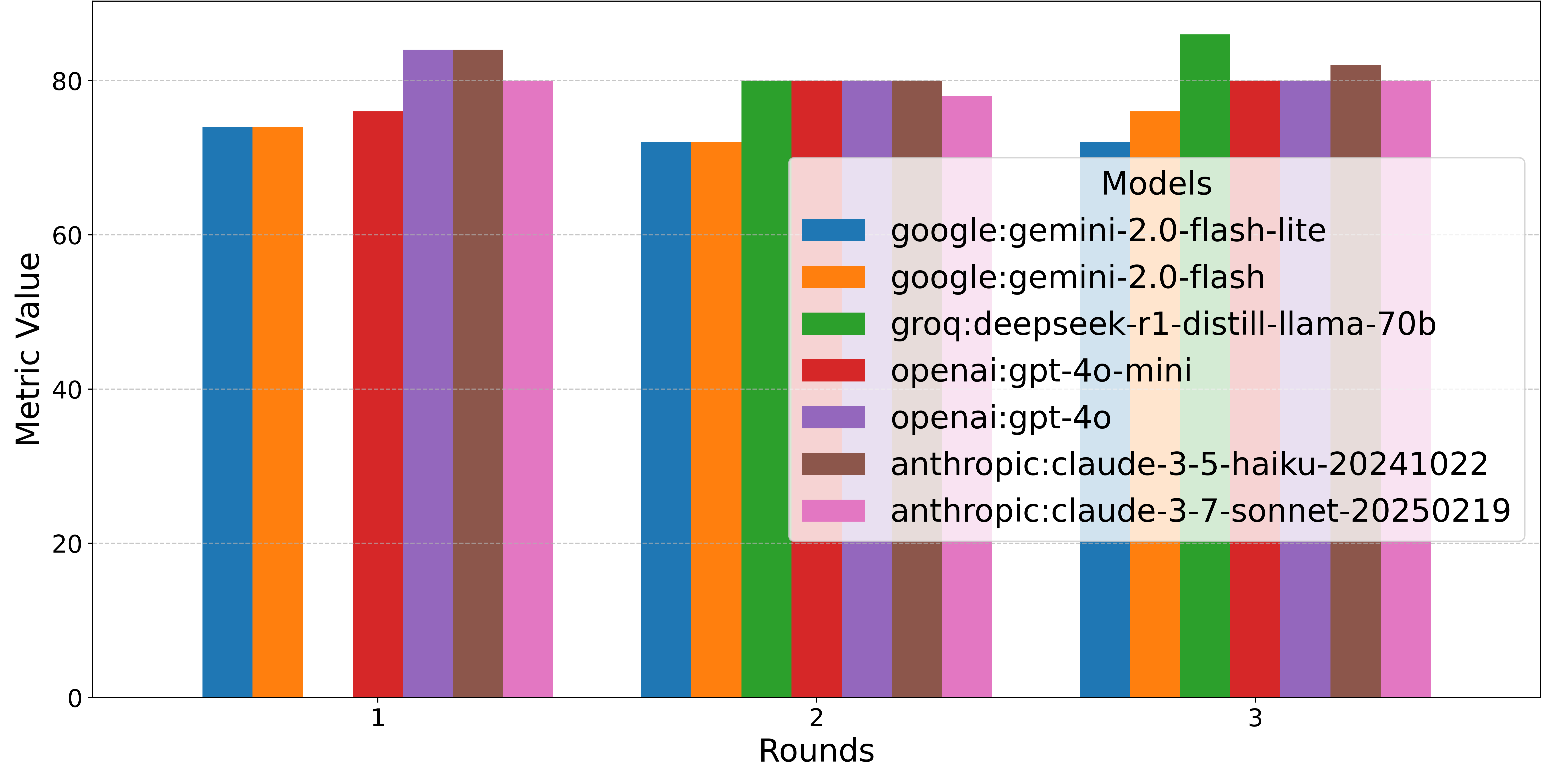}  
\caption{RAG answer match reference answer Conventional CTI Dataset}
\label{fig:rag_answer_match_reference_answer_dataset-security}
\end{figure}

\begin{figure}[th] 
\centering
\includegraphics[width=1\columnwidth]{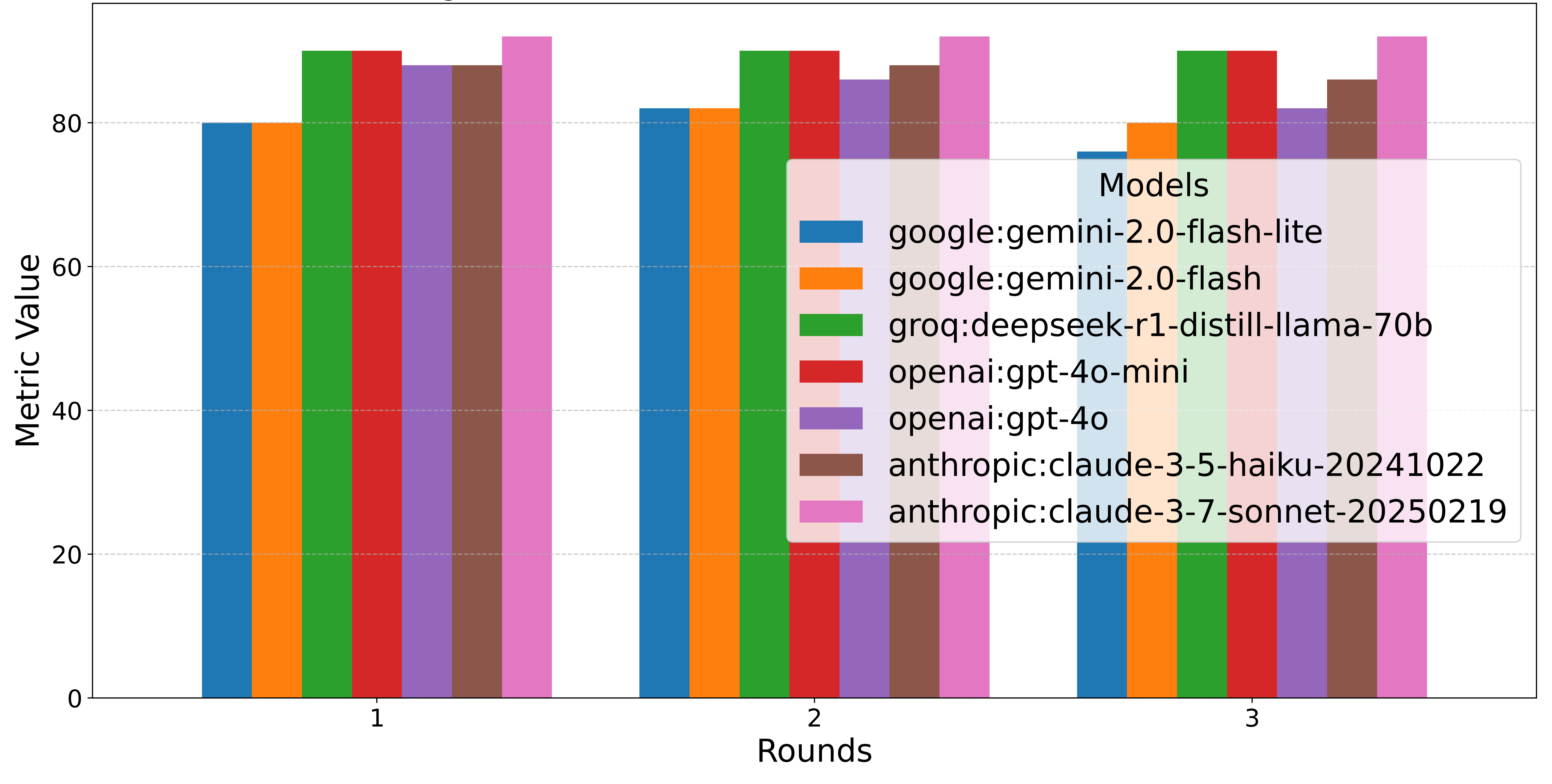}  
\caption{RAG answer match reference answer Blockchain CTI Dataset}
\label{fig:rag_answer_match_reference_answer_dataset-blockchain}
\end{figure}

The LLM self-evaluation showed high consistency, as evidenced by the ``RAG answer match Reference Answers" histograms for both datasets (Figures \ref{fig:rag_answer_match_reference_answer_dataset-security} and \ref{fig:rag_answer_match_reference_answer_dataset-blockchain}). However, slight variations in these histograms across the three test rounds, despite identical retrieved context for each query, suggested some variability in the generation or self-evaluation step.

This led to two primary hypotheses:

\begin{enumerate}
    \item The LLM self-evaluation occasionally made incorrect decisions about the semantic equivalence of an answer.
    \item The generation model sometimes overlooked relevant information within the provided context when formulating its response.
\end{enumerate}

To investigate this, a manual analysis of 2,050 LLM self-evaluation decisions was performed by reviewing testing logs and generated graphs. The verified correct answer percentages are presented in Table \ref{tab:performance_metrics_llm_self_evaluation_security_dataset} (Conventional CTI) and Table \ref{tab:performance_metrics_llm_self_evaluation_blockchain_dataset} (Blockchain CTI).

\begin{table}[htbp]
    \centering
    \caption{LLM Self Evaluation Performance Metrics Conventional CTI Dataset}
    \label{tab:performance_metrics_llm_self_evaluation_security_dataset}

    \begin{tabular}{llS[table-format=2.2]}
        \toprule
        \textbf{Name} & \textbf{Category} & {\textbf{Correct Decision}} \\ 
        \midrule
        Round 1 & Round & 93.00\% \\
        Round 2 & Round & 92.57\% \\
        Round 3 & Round & 93.43\% \\
        Google:gemini-2.0-flash-lite & Model & 92.67\% \\
        Google:gemini-2.0-flash & Model & 94.00\% \\
        Groq:deepseek-r1-distill-llama-70b & Model & 92.00\% \\
        Openai:gpt-4o-mini & Model & 93.33\% \\
        Openai:gpt-4o & Model & 94.67\% \\
        Anthropic:claude-3-5-haiku-20241022 & Model & 90.67\% \\
        Anthropic:claude-3-7-sonnet-20250219 & Model & 93.33\% \\
        \bottomrule
    \end{tabular}
\end{table}

\begin{table}[htbp]
    \centering
    \caption{LLM Self Evaluation Performance Metrics Blockchain CTI Dataset}
    \label{tab:performance_metrics_llm_self_evaluation_blockchain_dataset}

    \begin{tabular}{llS[table-format=2.2]}
        \toprule
        \textbf{Name} & \textbf{Category} & {\textbf{Correct Decision}} \\ 
        \midrule
        Round 1 & Round & 94.57\% \\
        Round 2 & Round & 94.57\% \\
        Round 3 & Round & 94.86\% \\
        Google:gemini-2.0-flash-lite & Model & 97.33\% \\
        Google:gemini-2.0-flash & Model & 96.00\% \\
        Groq:deepseek-r1-distill-llama-70b & Model & 92.00\% \\
        Openai:gpt-4o-mini & Model & 96.00\% \\
        Openai:gpt-4o & Model & 95.33\% \\
        Anthropic:claude-3-5-haiku-20241022 & Model & 91.33\% \\
        Anthropic:claude-3-7-sonnet-20250219 & Model & 94.67\% \\
        \bottomrule
    \end{tabular}
\end{table}

Our analysis yielded two key findings. First, a comparison of the RAG answers with the reference answers and the respective decision confirmed both of our initial hypotheses. Second, the locally hosted Mistral 7B model exhibited high consistency in its output across multiple runs during the LLM self-evaluation process. A slight performance advantage was observed on the Blockchain CTI dataset compared to the Conventional CTI dataset. Although this could suggest that its training data was more heavily weighted toward blockchain topics, it is more likely a result of the limited size of our experimental datasets. More testing with larger datasets is required to determine whether this performance difference is statistically significant.

\section{Conclusion}

This paper introduces RAGRecon, a system that integrates LLMs and RAG to efficiently process and visualize complex CTI. 

To evaluate our system, we benchmarked its performance on conventional and blockchain CTI datasets and demonstrated that it is possible to generate accurate, context-aware answers with high faithfulness. 

A key limitation we observed, particularly in models with up to 20 billion parameters, was their inconsistent reliability in processing data for the KG. Formatting errors were the primary cause of this problem. 

Despite this, our approach transforms the analysis of unstructured reports by not only automating the process but also allowing users to visualize threat relationships and derive clear explainable insights.

\footnotesize\vspace{5mm}\noindent\emph{Acknowledgments. } This work was financially supported by Project Blockchain.PT - Decentralize Portugal with Blockchain Agenda (Project no 51), WP6: Digital Assets Management, Call no 02/C05-i01.01/2022, funded by the Portuguese Recovery and Resilience Program (PPR), The Portuguese Republic and The European Union (EU) under the framework of Next Generation EU Program. This work was also supported by national funds through Funda\c{c}\~{a}o para a Ci\^{e}ncia e a Tecnologia (FCT) with reference UIDB/50021/2020 (INESC-ID).

\bibliographystyle{IEEEtran}
\bibliography{./Bibliography}

@article{zhang2024llms,
  title={When {LLMs} meet cybersecurity: A systematic literature review},
  author={Zhang, Jie and Bu, Haoyu and Wen, Hui and Chen, Yu and Li, Lun and Zhu, Hongsong},
  journal={arXiv preprint arXiv:2405.03644},
  year={2024}
}

@article{roffo2024exploring,
  title={Exploring Advanced Large Language Models with {LLMsuite}},
  author={Roffo, Giorgio},
  journal={arXiv preprint arXiv:2407.12036},
  year={2024}
}

@article{lewis2020retrieval,
  title={Retrieval-augmented generation for knowledge-intensive {NLP} tasks},
  author={Lewis, Patrick and Perez, Ethan and Piktus, Aleksandra and Petroni, Fabio and Karpukhin, Vladimir and Goyal, Naman and K{\"u}ttler, Heinrich and Lewis, Mike and Yih, Wen-tau and Rockt{\"a}schel, Tim and others},
  journal={Advances in Neural Information Processing Systems},
  volume={33},
  pages={9459--9474},
  year={2020}
}

@article{siracusano2023time,
  title={Time for action: Automated analysis of cyber threat intelligence in the wild},
  author={Siracusano, Giuseppe and Sanvito, Davide and Gonzalez, Roberto and Srinivasan, Manikantan and Kamatchi, Sivakaman and Takahashi, Wataru and Kawakita, Masaru and Kakumaru, Takahiro and Bifulco, Roberto},
  journal={arXiv preprint arXiv:2307.10214},
  year={2023}
}

@article{martins2022generating,
  title={Generating quality threat intelligence leveraging {OSINT} and a cyber threat unified taxonomy},
  author={Martins, Cl{\'a}udio and Medeiros, Ib{\'e}ria},
  journal={ACM Transactions on Privacy and Security},
  volume={25},
  number={3},
  pages={1--39},
  year={2022},
  publisher={ACM New York, NY}
}

@article{alves2021processing,
  title={Processing tweets for cybersecurity threat awareness},
  author={Alves, Fernando and Bettini, Aur{\'e}lien and Ferreira, Pedro M and Bessani, Alysson},
  journal={Information Systems},
  volume={95},
  pages={101586},
  year={2021},
  publisher={Elsevier}
}

@article{shafee2024evaluation,
  title={Evaluation of {LLM}-based chatbots for {OSINT}-based Cyber Threat Awareness},
  author={Shafee, Samaneh and Bessani, Alysson and Ferreira, Pedro M},
  journal={Expert Systems with Applications},
  pages={125509},
  year={2024},
  publisher={Elsevier}
}

@article{rajapaksha2024rag,
  title={A {RAG}-Based Question-Answering Solution for Cyber-Attack Investigation and Attribution},
  author={Rajapaksha, Sampath and Rani, Ruby and Karafili, Erisa},
  journal={arXiv preprint arXiv:2408.06272},
  year={2024}
}

@article{mitra2024localintel,
  title={Localintel: Generating organizational threat intelligence from global and local cyber knowledge},
  author={Mitra, Shaswata and Neupane, Subash and Chakraborty, Trisha and Mittal, Sudip and Piplai, Aritran and Gaur, Manas and Rahimi, Shahram},
  journal={arXiv preprint arXiv:2401.10036},
  year={2024}
}

@inproceedings{es2024ragas,
  title={{RAGAS}: Automated evaluation of retrieval augmented generation},
  author={Es, Shahul and James, Jithin and Anke, Luis Espinosa and Schockaert, Steven},
  booktitle={Proceedings of the 18th Conference of the European Chapter of the Association for Computational Linguistics: System Demonstrations},
  pages={150--158},
  year={2024}
}

@article{fayyazi2023advancing,
  title={Advancing {TTP} Analysis: Harnessing the Power of Encoder-Only and Decoder-Only Language Models with Retrieval Augmented Generation},
  author={Fayyazi, Reza and Taghdimi, Rozhina and Yang, Shanchieh Jay},
  journal={arXiv preprint arXiv:2401.00280},
  year={2023}
}

@article{liu2023lostmiddlelanguagemodels,
  title={Lost in the middle: How language models use long contexts},
  author={Liu, Nelson F and Lin, Kevin and Hewitt, John and Paranjape, Ashwin and Bevilacqua, Michele and Petroni, Fabio and Liang, Percy},
  journal={Transactions of the Association for Computational Linguistics},
  volume={12},
  pages={157--173},
  year={2024},
  publisher={MIT Press One Broadway, 12th Floor, Cambridge, Massachusetts 02142, USA~…}
}

@article{hornuf2023cybercrime,
  title={Cybercrime on the {Ethereum} blockchain},
  author={Hornuf, Lars and Momtaz, Paul P and Nam, Rachel J and Yuan, Ye},
  journal={CESifo Working Paper No. 10598},
  year={2023},
}

@article{cong2023blockchain,
  title={Blockchain forensics and crypto-related cybercrimes},
  author={Cong, Lin William and Grauer, Kimberly and Rabetti, Daniel and Updegrave, Henry},
  journal={Available at SSRN 4358561},
  year={2023}
}

@article{dahl2024large,
  title={Large legal fictions: Profiling legal hallucinations in large language models},
  author={Dahl, Matthew and Magesh, Varun and Suzgun, Mirac and Ho, Daniel E},
  journal={Journal of Legal Analysis},
  volume={16},
  number={1},
  pages={64--93},
  year={2024},
  publisher={Oxford University Press UK}
}

@article{pal2023med,
  title={Med-halt: Medical domain hallucination test for large language models},
  author={Pal, Ankit and Umapathi, Logesh Kumar and Sankarasubbu, Malaikannan},
  journal={arXiv preprint arXiv:2307.15343},
  year={2023}
}

@article{yenduri2024gpt,
  title={{GPT} (generative pre-trained transformer)--a comprehensive review on enabling technologies, potential applications, emerging challenges, and future directions},
  author={Yenduri and others},
  journal={IEEE Access},
  year={2024},
  publisher={IEEE}
}

@inproceedings{ren2023self,
  title={Self-evaluation improves selective generation in large language models},
  author={Ren, Jie and Zhao, Yao and Vu, Tu and Liu, Peter J and Lakshminarayanan, Balaji},
  booktitle={Proceedings on "I Can't Believe It's Not Better: Failure Modes in the Age of Foundation Models" at NeurIPS 2023 Workshops},
  pages={49--64},
  year={2023},
}

@article{schneider2024explainable,
  title={{Explainable Generative AI} ({GenXAI}): A survey, conceptualization, and research agenda},
  author={Schneider, Johannes},
  journal={Artificial Intelligence Review},
  volume={57},
  number={11},
  pages={289},
  year={2024},
  publisher={Springer}
}

@article{rajabi2022knowledge,
  title={{Knowledge Graphs and Explainable AI in Healthcare}},
  author={Rajabi, Enayat and Kafaie, Somayeh},
  journal={Information},
  volume={13},
  number={10},
  pages={459},
  year={2022},
  publisher={MDPI}
}

@article{minaee2024large,
  title={{Large Language Models: A Survey}},
  author={Minaee, Shervin and Mikolov, Tomas and Nikzad, Narjes and Chenaghlu, Meysam and Socher, Richard and Amatriain, Xavier and Gao, Jianfeng},
  journal={arXiv preprint arXiv:2402.06196},
  year={2024}
}

@article{arrieta2020explainable,
  title={Explainable Artificial Intelligence (XAI): Concepts, taxonomies, opportunities and challenges toward responsible AI},
  author={Arrieta, Alejandro Barredo and D{\'\i}az-Rodr{\'\i}guez, Natalia and Del Ser, Javier and Bennetot, Adrien and Tabik, Siham and Barbado, Alberto and Garc{\'\i}a, Salvador and Gil-L{\'o}pez, Sergio and Molina, Daniel and Benjamins, Richard and others},
  journal={Information fusion},
  volume={58},
  pages={82--115},
  year={2020},
  publisher={Elsevier}
}

@article{goodfellow2020generative,
  title={Generative adversarial networks},
  author={Goodfellow, Ian and Pouget-Abadie, Jean and Mirza, Mehdi and Xu, Bing and Warde-Farley, David and Ozair, Sherjil and Courville, Aaron and Bengio, Yoshua},
  journal={Communications of the ACM},
  volume={63},
  number={11},
  pages={139--144},
  year={2020},
  publisher={ACM New York, NY, USA}
}

@article{naveed2023comprehensive,
  title={A comprehensive overview of large language models},
  author={Naveed, Humza and Khan, Asad Ullah and Qiu, Shi and Saqib, Muhammad and Anwar, Saeed and Usman, Muhammad and Akhtar, Naveed and Barnes, Nick and Mian, Ajmal},
  journal={ACM Transactions on Intelligent Systems and Technology},
  year={2023},
  publisher={ACM New York, NY}
}

@inproceedings{devlin2019bert,
  title={Bert: Pre-training of deep bidirectional transformers for language understanding},
  author={Devlin, Jacob and Chang, Ming-Wei and Lee, Kenton and Toutanova, Kristina},
  booktitle={Proceedings of the 2019 conference of the North American chapter of the association for computational linguistics: human language technologies, volume 1 (long and short papers)},
  pages={4171--4186},
  year={2019}
}

@article{achiam2023gpt,
  title={Gpt-4 technical report},
  author={Achiam, Josh and Adler, Steven and Agarwal, Sandhini and Ahmad, Lama and Akkaya, Ilge and Aleman, Florencia Leoni and Almeida, Diogo and Altenschmidt, Janko and Altman, Sam and Anadkat, Shyamal and others},
  journal={arXiv preprint arXiv:2303.08774},
  year={2023}
}

\end{document}